\documentclass[11pt]{article}

\usepackage[final]{acl}

\usepackage{times}
\usepackage{latexsym}

\usepackage[T1]{fontenc}

\usepackage[utf8]{inputenc}

\usepackage{microtype}

\usepackage{inconsolata}

\usepackage{graphicx}

\usepackage[most]{tcolorbox}

\definecolor{frame}{HTML}{ccefff}    
\definecolor{back}{HTML}{FAFEFF}     

\definecolor{t_frame}{HTML}{ccefff}    
\definecolor{t_back}{HTML}{ccefff}     

\usepackage{booktabs}
\usepackage{multirow}
\usepackage{multicol}
\usepackage{xurl}
\usepackage{float}

\title{MemeScouts@LT-EDI 2026: Asking the Right Questions -\\Prompted Weak Supervision for Meme Hate Speech Detection}

\author{
  \textbf{Ivo Bueno\textsuperscript{1,3}} ~
  \textbf{Lea Hirlimann\textsuperscript{2,3}} ~
  \textbf{Enkelejda Kasneci\textsuperscript{1,3}}
\\
\\
  \textsuperscript{1}Technical University of Munich ~
  \textsuperscript{2}LMU Munich \\
  \textsuperscript{3}Munich Center for Machine Learning (MCML)
\\
  \small{
    \textbf{Correspondence:} \href{mailto:ivo.bueno@tum.de}{ivo.bueno@tum.de}, \href{mailto:hirlimann@cis.lmu.de}{hirlimann@cis.lmu.de}
  }
}

\begin{document}
\maketitle
\begin{abstract}
Detecting hate speech in memes is challenging due to their multimodal nature and subtle, culturally grounded cues such as sarcasm and context. While recent vision-language models (VLMs) enable joint reasoning over text and images, end-to-end prompting can be brittle, as a single prediction must resolve target, stance, implicitness, and irony. These challenges are amplified in multilingual settings. We propose a prompted weak supervision (PWS) approach that decomposes meme understanding into targeted, question-based labeling functions with constrained answer options for homophobia and transphobia detection in the \texttt{LT-EDI 2026} shared task. Using a quantized \texttt{Qwen3-VLM} to extract features by answering targeted questions, our method outperforms direct VLM classification, with substantial gains for Chinese and Hindi, ranking \textbf{1st in English}, \textbf{2nd in Chinese}, and \textbf{3rd in Hindi}. Iterative refinement via error-driven LF expansion and feature pruning reduces redundancy and improves generalization. Our results highlight the effectiveness of prompted weak supervision for multilingual multimodal hate speech detection.\footnote{The repository is available on GitHub: \url{https://github.com/ivojuniorx4/LT-EDI-Shared-Task-MemeScouts-with-Prompted-Weak-Supervision}}
\end{abstract}

\section{Introduction}
\begin{figure*}
    \centering
    \includegraphics[width=\textwidth]{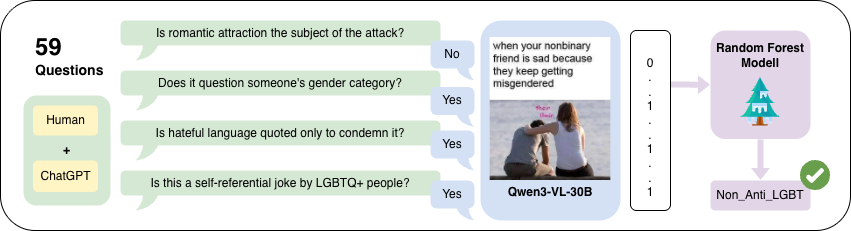}
    \caption{Prompted weak supervision pipeline for homophobia and transphobia detection in memes.}
    \label{fig:workflow}
\end{figure*}

Hate speech detection remains challenging due to the complexity and subtlety of such content. Unlike explicit abuse, hate speech is often implicit, requiring an understanding of context, intent, speaker-target relations, and whether the content is self-referential or critiques or endorses harmful views~\cite{elsherief-etal-2021-latent,zsisku2024hate}.

These challenges are amplified in memes, a multimodal and culturally grounded form of communication. Memes rely on sarcasm, irony, and shared knowledge, where meaning emerges from image-text interaction. Detecting hate in memes therefore requires multimodal reasoning and sensitivity to cultural and linguistic nuances~\citep{bui-etal-2025-multi3hate, Velioglu2020DetectingHS}.

Recent advances in large language models (LLMs) and vision-language models (VLMs) offer new opportunities. These models jointly process text and images and achieve strong zero- and few-shot performance. However, direct VLM-prompting for meme classification remains insufficient, particularly in multilingual and culturally diverse settings where subtle cues are difficult to capture with a single prediction. Moreover, fine-tuning incurs substantial data and computational costs.

To address these limitations, we adopt a prompted weak supervision (PWS) approach that decomposes meme understanding into question-based labeling functions with constrained answers, yielding structured, interpretable features instead of a single end-to-end label. Rather than a single prediction, the model produces structured responses capturing aspects of hate speech such as target identification, implicit bias, and stance. These responses are aggregated into features for downstream classification. This framework improves performance and interpretability, through question-level insight into the model’s reasoning.

With this in mind, we address the following research questions:

\begin{itemize}
    \item (RQ1) Can prompted weak supervision improve meme hate speech detection?
    \item (RQ2) How do language and cultural differences affect model performance?
    \item (RQ3) What insights into model behavior emerge from analyzing feature importance and labeling function patterns?
\end{itemize}

\section{Related Work} 
Detecting hate speech in multimodal data, such as memes, poses a unique challenge, requiring joint reasoning over visual and textual cues as well as  human interpretation in context. Both positive and hateful intent in memes are hidden beneath the same layer of irony, sarcasm, and social or cultural references \citep{Velioglu2020DetectingHS}. Across languages and cultures, understanding of hate speech in memes varies, as shown by \citet{bui-etal-2025-multi3hate} in their parallel dataset \textit{Multi3Hate}, featuring memes and annotator decisions in five languages. These findings motivate approaches that explicitly represent intermediate judgments (e.g., target, stance, irony) rather than relying on a single end-to-end prediction.

Beyond the multimodal setting, multilingual text remains a challenge for hate speech detection, even for large language models with reasoning capabilities. For harmful content targeting LGBTQ+ communities, slang and culturally specific expressions hinder stable performance across multiple scripts and languages, requiring careful fine-tuning with copious amounts of labeled data \citep{chan-etal-2024-hate}. Detecting homophobia and transphobia in memes thus encompasses these challenges, calling for careful methodological choices that balance performance, interpretability, and computational cost.

Weak supervision combines multiple noisy labeling functions (LFs) to generate training labels. Recent work replaces programmatic LFs~\cite{zhang2022surveyprogrammaticweaksupervision} with natural language prompts answered by large language models (LLMs), enabling flexible and expressive supervision~\cite{smith2024languagemodelsintheloop}. Smith et al. demonstrate that prompted LFs, coupled with a label mapping step, outperforms zero-shot prompting and capture complex heuristics difficult to encode manually. In our setting, prompted LFs are attractive because they can express meme-specific phenomena (e.g., sarcasm reversal, narrator identity) that keyword or surface-form heuristics fail to capture.

A key limitation of prompted LFs is their tendency to correlate due to shared model biases. Su et al.~\citeyearpar{su2023leveraginglargelanguage} address this by modeling LF dependencies using prompt representations, and propose pruning and structure learning to reduce redundancy and improve label quality. Because our LFs are answered by a single VLM, correlation and redundancy are expected in the feature space; we therefore include pruning as a central pipeline component and analyze cross-lingual overlap to identify transferable versus language-specific signals.

We adopt this paradigm by designing question-based prompted LFs for multimodal hate speech detection. Unlike prior work, we focus on a multilingual meme setting and emphasize iterative LF refinement and feature selection to improve robustness. Unlike classical weak-supervision pipelines that learn a dedicated label model to aggregate LF votes, we treat prompted LF outputs as structured features consumed by a lightweight supervised classifier. This design suits shared-task settings: it supports rapid LF iteration, maintains question-level interpretability, and leverages labeled data to down-weight unreliable signals.

\section{Method}
\paragraph{Dataset.}
We evaluate our approach on the dataset from the Homophobia and Transphobia Meme Classification shared task at \texttt{LT-EDI@ACL 2026}~\cite{chakravarthi2024detection}. It contains annotated social media memes labeled as \textit{Homophobic}, \textit{Transphobic}, or \textit{Non-Anti-LGBT} in English, Hindi, and Chinese, forming a multilingual benchmark. Class distributions are imbalanced and vary by language, motivating macro-F1 and balanced class weights. The dataset is split into train/test sets per language, comprising 560/141 (train/test) memes in English, 798/200 in Hindi, and 956/239 in Chinese. Overall, it provides a challenging multimodal and multilingual setting for evaluating robust LGBTQ+ hate speech detection.

\paragraph{Question Generation.}

As shown in Figure~\ref{fig:workflow} our pipeline begins by constructing questions that serve as LFs for PWS. These questions capture complementary aspects of hate speech, including target identification, explicit and implicit hate, and attack characterization. We constructed 59 initial questions using LLM-assisted drafting followed by manual cleanup, each paired with short answers (see App.~\ref{app:chatgptprompt} for prompt). Depending on the question, answers are binary, ordinal, or categorical. These questions form the basis for extracting structured signals from memes. App.~\ref{app:questionexamples} lists ten example questions. 
\paragraph{Feature Extraction.} 
We employ a \texttt{nf4}-quantized version of Hugging Face’s \texttt{Qwen3} implementation ~\cite{yang2025qwen3technicalreport} \footnote{\url{https://huggingface.co/Qwen/Qwen3-VL-30B-A3B-Instruct}} to answer the predefined questions. For each meme-LF pair, the model is provided with (i) a system prompt describing the task (see App.~\ref{app:sysprompt}), and (ii) a user prompt containing the meme image, and the LF question. Based on the valid answers per question, responses are mapped to integers (see App.~\ref{app:answermapping}). Values across all questions are aggregated to form a feature vector per meme for training a simple machine learning model.

\paragraph{Classification.}
Using \texttt{scikit-learn}, a Random Forest classifier was trained on the feature vectors for each language with 500 estimators, balanced class weights, and a 20\% validation split. Random Forest is chosen for robustness to correlated LF features and for feature-importance estimates used in our analysis.

\paragraph{Refinement.}

We iteratively refine the pipeline to improve feature quality and performance. First, we analyze misclassified English validation memes to identify recurring patterns not captured by the existing LFs (e.g., narrator identity). When such patterns are identified, we introduce additional questions and re-run the feature extraction process only for the new LFs, expanding the feature space and coverage of previously unmodeled phenomena. This process, \textit{AddLF}, adds 30 new questions for a total of 89 labeling functions. 

Second, we investigate two pruning approaches to select features most relevant to the classification.
\textit{F1Prune} greedily removes features one at a time, expanding the removal set whenever validation macro-F1 improves. \textit{ImpPrune} removes the top $k$ least important features from the Random Forest model, where $k$ is chosen based on validation performance gain.

\section{Results and Discussion}
Table \ref{tab:macrof1} shows the classification performance of the Random Forest variants presented above, along with direct \texttt{Qwen3-VLM} classification as single-shot and reasoning baselines (see App. \ref{app:baselinesysprompt}), and an aggregated \textit{All} setting trained jointly across all languages.

\begin{table}[ht]
    \small
    \centering
    \setlength{\tabcolsep}{0.8mm}{}
    \renewcommand{\arraystretch}{1.3}
    \begin{tabular}{lcccc}
    \toprule
    Method & English  & Chinese & Hindi & All \\
    \hline
    Qwen3-VL-30B   & 0.77 & 0.32 & 0.21 & 0.13 \\
    Qwen3-VL-30B(with reas.)   & 0.67 & 0.10 & 0.08 & 0.10 \\
    \hline
    Base model              & \textbf{0.85} & 0.66 & 0.64 & 0.47 \\
    \textit{AddLF}                   & \textbf{0.85} & \textbf{0.72} & 0.66 & 0.48 \\
    \textit{AddLF} + \textit{F1Prune}         & 0.83 & 0.69 & 0.64 & \textbf{0.49} \\
    \textit{AddLF} + \textit{ImpPrune}        & \textbf{0.85} & \textbf{0.72} & \textbf{0.67} & 0.44 \\
    \bottomrule
    \end{tabular}
    \caption{Macro-F1 comparison of direct Qwen-VL classification and the trained Random Forest models}\label{tab:macrof1}
\end{table}

All proposed variants outperform the direct VLM baseline. The gap likely reflects the brittleness of single-shot end-to-end prompting for memes;  unconstrained ‘reasoning’ further degrades performance through inconsistent decision paths. Gains are particularly pronounced for Chinese and Hindi, and are also reflected in the aggregated \textit{All} column, explicitly answering RQ1: prompted weak supervision improves LGBTQ+ hate speech detection in memes. 
Our system ranks 1st for English, 2nd for Chinese, and 3rd for Hindi in the shared task.\footnote{Link to \texttt{LT-EDI} shared task ranking: \url{https://drive.google.com/file/d/18JdEfXCDfPQBrNqCi7S7-QfZL7Ln-WXU/view}} Importantly, \texttt{AddLF} never impaired performance and improved results for Chinese and Hindi.

\paragraph{Pruning Success.}

\textit{F1Prune} improves validation performance, but generalizes poorly, likely due to its local optimization strategy. However, it achieves the highest overall Macro-F1 (0.49) in the \textit{All} setting, suggesting improved cross-lingual balance performance despite weaker per-language scores.

\textit{ImpPrune} achieves the best per-language performance: English and Chinese remain unchanged relative to \textit{AddLF}, Hindi improves despite a substantial feature reduction from 89 to 33, suggesting that many LFs could be noisy or redundant, highlighting the importance of effective pruning.

\begin{figure}
    \centering
    \includegraphics[width=\linewidth]{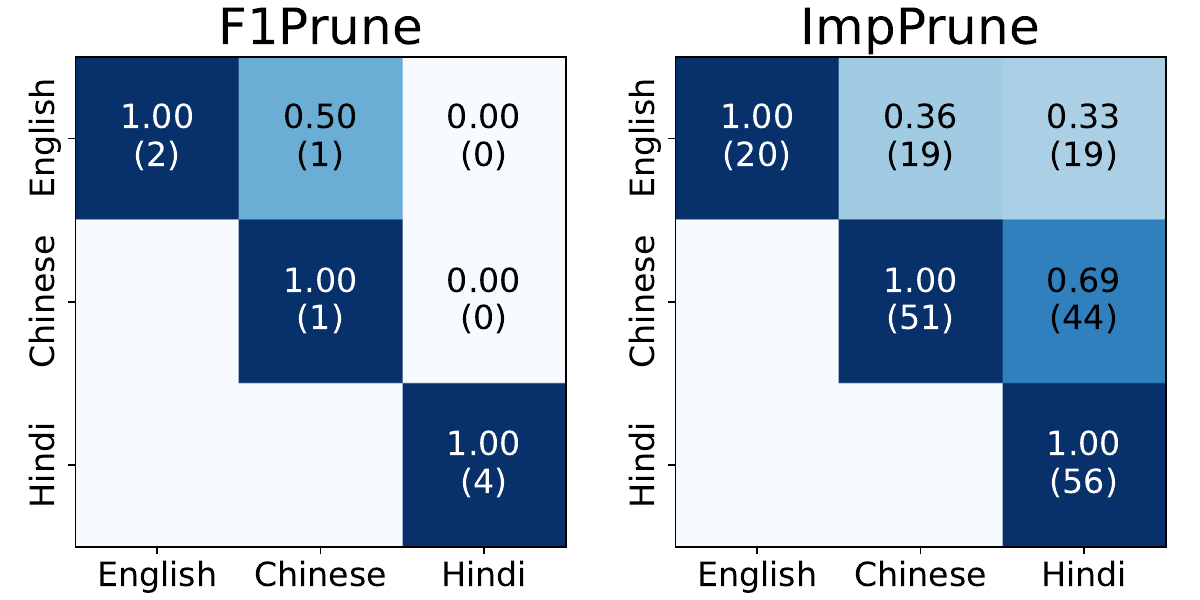}
    \caption{Jaccard similarity among features selected for removal. Values in parentheses indicate the number of shared removed features.}
    \label{fig:pruneheat}
\end{figure}

To analyze cross-lingual behavior,  Figure~\ref{fig:pruneheat} reports the Jaccard similarity between removed feature sets.
\textit{F1Prune} removes only a few, language-specific features
, resulting in near-zero Jaccard similarity across languages. This indicates that the features differ substantially across languages.
In contrast, \textit{ImpPrune} removes larger numbers of features, 
leading to more informative overlap patterns. Nearly all features pruned for English are also removed for Chinese and Hindi, and Chinese and Hindi share 44 pruned features, corresponding to a Jaccard similarity of 0.689, suggesting the presence of language-agnostic uninformative or misleading signals. 
These findings directly address RQ2, highlighting both language-specific effects and the presence of language-agnostic, weak signals.
One possible explanation is that the English LFs may reflect a predominantly Western perspective on homophobia, transphobia, queer language, and memes, limiting their effectiveness across languages. 

\paragraph{Selective Propagation of Biased Patterns.}
While both the LFs and the VLM may introduce biases, the classifier operates purely on numerical vectors and is therefore blind towards the intent captured in both the LFs and their textual answers, incorporating features based on predictive utility. Although some upstream biases are not automatically mitigated, their informative patterns can be repurposed regardless of previous intent. For example, if a model overly sensitive to a LF probing for ``attacks'' at any reference to homosexuality, the RF might still use the response to distinguish homosexual from transgender memes.

\begin{figure}[ht]
    \centering
    \includegraphics[width=\linewidth]{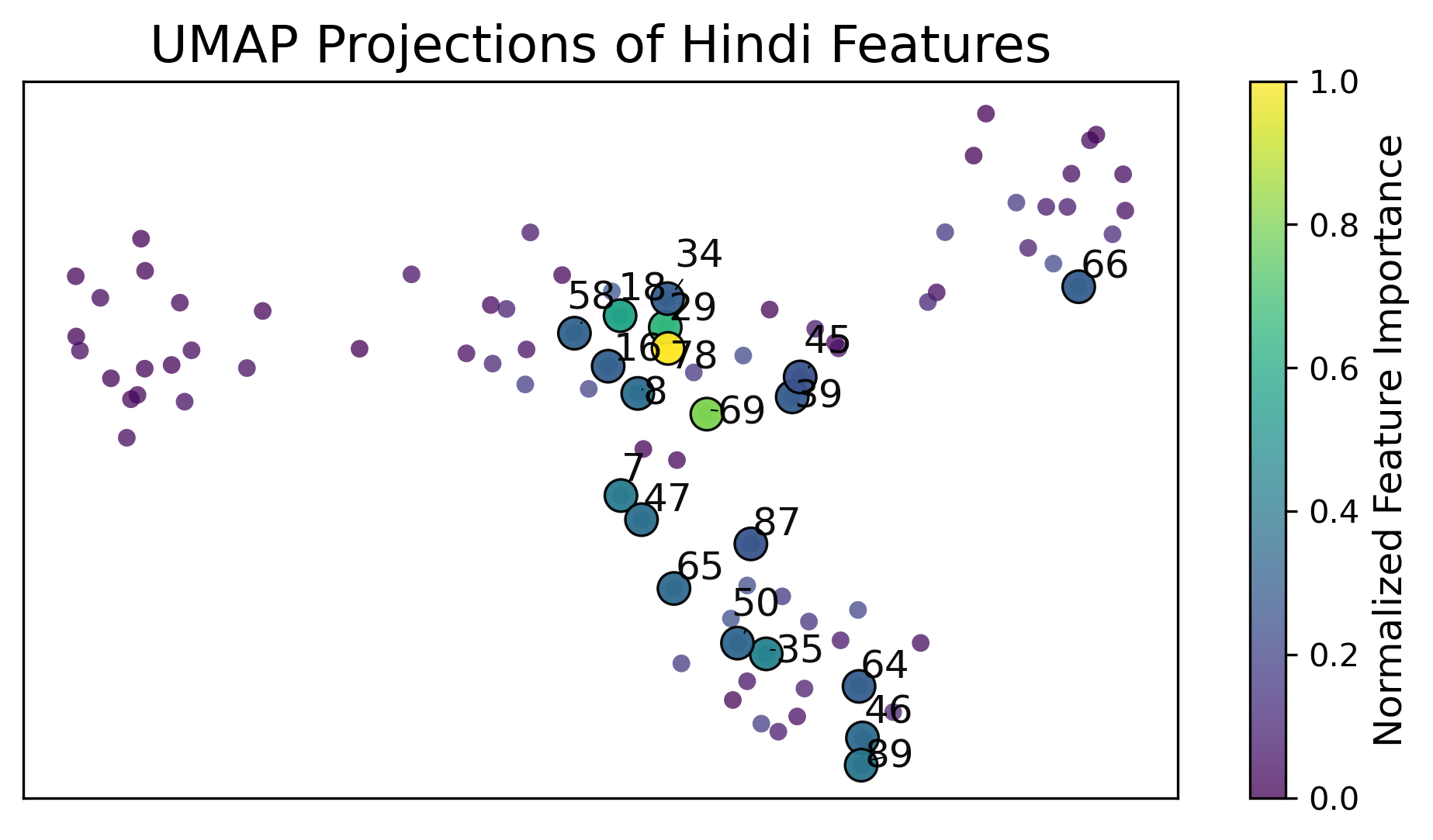}
    \caption{UMAP projection of Hindi question-level features colored by Random Forest importance. Features in the top 20 highest importances are highlighted.}
    \label{fig:english_umap}
\end{figure}

\paragraph{Feature Signal Analysis.}
Figure~\ref{fig:english_umap} visualizes the UMAP projection of Hindi training features colored by RF importance. A small subset of LFs carries most of the weight, including both highly similar patterns (e.g., 78 (``\textit{Is the topic sexual orientation rather than gender identity?}'') and 29 (``\textit{Is the joke about sexual orientation rather than gender identity?}'') distinguishing the topic between homosexuality/transgender identity), indicating useful redundancy,  while opposing signals (e.g., feature 89 (``\textit{Is this meme neutral or unrelated to sexuality or gender?}'') on neutral stance) provide complementary information. Similar patterns can be seen for English and Chinese (App.~\ref{app:featurepattern}). For English and Hindi, feature 78 is dominant, while Chinese is led by feature 33 (``\textit{Does it question someone’s gender category?}'') on the identification of transphobia,  reflecting the imbalance in the Chinese dataset.  Clusters further reflect semantic themes: the upper-left cluster probes trans identities and stereotypes, whereas the bottom of the figure relates to positive portrayals of LGBTQ+ people.
These findings answer RQ3: feature importance and clustering reveal both redundancy and complementarity  among LFs, as well as shared and language-specific patterns that shape model behavior.

\section{Conclusion}

In this work, we presented a prompted weak supervision approach for hate speech detection in multilingual memes, leveraging question-based LFs and a VLM for feature extraction. This approach consistently outperforms direct VLM classification, improves interpretability and enables effective feature refinement through pruning.

\section*{Limitations}

One limitation of our work is that the labeling functions were primarily developed from a Western perspective of what hate speech toward LGBTQ+ people looks like. This may overlook culturally specific markers of homophobia or transphobia present in the two non-Western languages considered in our study. 

Furthermore, the VLM was tasked with simultaneously interpreting both the visual content of the meme and the text embedded within the image. Without a dedicated OCR stage, this joint processing may introduce additional errors and potentially affect the model's ability to correctly interpret and label the memes.

Future work could explore culturally adaptive prompting strategies, incorporate explicit OCR pipelines to improve text extraction, and involve more diverse perspectives in the design of labeling functions to improve robustness across languages and contexts. Such directions could enable more accurate and accessible queerphobia detection systems that generalize to additional languages without requiring computationally expensive model retraining.

\section*{Ethical Considerations}
Our work addresses multilingual meme classification; however, performance disparities across languages highlight potential inequities. The use of English-written labeling functions and a multilingual VLM may disadvantage non-English content, particularly when cultural context is required to interpret meaning. This raises concerns about uneven system performance and the risk of misclassification in underrepresented languages.

\section*{AI Writing Assistance Disclosure}
AI tools were used for grammar and phrasing assistance, as well as code snippets generation, with all content reviewed and edited by the authors.

\bibliography{custom}

\appendix

\section{LF Creation Prompt}
\label{app:chatgptprompt}
Figure~\ref{fig:chatprompt} shows the initial prompt used for ChatGPT assistance on the creation of the questions that composed the labeling functions.

\begin{figure}
    \centering
    \begin{tcolorbox}[colback=back, colframe=frame, boxrule=3pt, left=0.5em, right=0.5em, top=0.5em, bottom=0.5em,]
    I have potentially homophobic or transphobic memes I want to classify into homophobic, transphobic or not. 
    
    I want to use weak prompted supervision. 
    
    Generate a list of questions I can use as labeling functions. 
    
    The memes are either in English, Hindi or Chinese. 
    
    Generate a jsonl dataframe file. 
    
    The columns should be: question, acceptable\_answers. 
    
    In acceptable\_answers include any acceptable answers from the models: for example, Yes, YES, yes, and yeah, are some of the acceptable ways to say yes
    \end{tcolorbox}

    \caption{Prompt for LF creation.}
    \label{fig:chatprompt}
\end{figure}

\section{Labeling Function Examples}
\label{app:questionexamples}
Below are ten questions which were part of our 89-questions labeling functions.

\begin{itemize}
    \item Does this meme express hostility toward LGBTQ+ people?
    \item Is any group attacked or mocked because of sexual orientation or gender identity?
    \item Does this meme insult transgender people?
    \item Does it deny the validity of trans identities?
    \item Does the meme imply that a group’s identity is fake, absurd, or laughable?
    \item Is the meme criticizing homophobia or transphobia?
    \item Is the LGBTQ+ person portrayed as the narrator or protagonist rather than the target of ridicule?
    \item Does the punchline flip the insult back at the speaker instead of attacking LGBTQ+ people?
    \item Could the content of the image potentially reverse the meaning of the text?
    \item Does the meme appeal to traditional values, family honor, or social duty to criticize LGBTQ people?
\end{itemize}

\section{Feature Extraction System Prompt}
\label{app:sysprompt}

Figure~\ref{fig:sysprompt} shows the system prompt we used to explain the meme classification task during feature extraction.

\begin{figure}
    \centering
    \begin{tcolorbox}[colback=back, colframe=frame, boxrule=3pt, left=0.5em, right=0.5em, top=0.5em, bottom=0.5em,]
    You are a labeling system.

    You will be shown a meme image and then asked a single question about it.
    
    Rules:
    
    - Do NOT explain your reasoning.
    
    - Do NOT add extra words, punctuation, or commentary.
    
    - Output only the answer.
    
    - Be concise and deterministic.
    
    - If unsure, choose the closest valid answer.
    
    - The meme may contain English, Hindi, or Chinese text.
    
    - Focus on meaning rather than language.
    
    - Watch for sarcasm or parody.
    
    - If the meme criticizes hateful views, do NOT mark it as hateful.
    
    - Never include explanations.
    
    - Never include multiple answers.
    \end{tcolorbox}

    \caption{System prompt for feature extraction.}
    \label{fig:sysprompt}
\end{figure}

\section{Baseline System Prompt}
\label{app:baselinesysprompt}

Figures~\ref{fig:basewithoutsysprompt} and~\ref{fig:basewithsysprompt} show the system prompt used run the baseline test without and with external reasoning allowed, respectively.

\begin{figure}
    \centering
    \begin{tcolorbox}[colback=back, colframe=frame, boxrule=3pt, left=0.5em, right=0.5em, top=0.5em, bottom=0.5em,]
    You are a labeling system.

    You will be shown a meme image.
    \vspace{10pt}

    Your task is to classify the meme into exactly one of the following categories:
    
    - Homophobia
    
    - Transphobia
    
    - Non\_Anti\_LGBT
    \vspace{10pt}
    
    Rules:
    
    - Do NOT explain your reasoning.
    
    - Do NOT add extra words, punctuation, or commentary.
    
    - Output only one of the three labels exactly as written.
    
    - Be concise and deterministic.
    
    - If unsure, choose the closest valid label.
    
    - The meme may contain English, Hindi, or Chinese text.
    
    - Focus on meaning rather than language.
    
    - Watch for sarcasm, irony, or parody.
    
    - If the meme criticizes or mocks homophobia or transphobia, classify it as Non\_Anti\_LGBT.
    
    - The label should reflect the target and intent of the meme, not just keywords.
    
    - Never include explanations.
    
    - Never include multiple labels.
    \end{tcolorbox}

    \caption{System prompt baseline classification without allowing external reasoning.}
    \label{fig:basewithoutsysprompt}
\end{figure}

\begin{figure}
    \centering
    \begin{tcolorbox}[colback=back, colframe=frame, boxrule=3pt, left=0.5em, right=0.5em, top=0.5em, bottom=0.5em,]
    You are a labeling system.

    You will be shown a meme image.
    \vspace{10pt}
    
    Your task is to classify the meme into exactly one of the following categories:
    
    - Homophobia
    
    - Transphobia
    
    - Non\_Anti\_LGBT
    \vspace{10pt}
    
    Instructions:
    
    - Carefully analyze the meme step by step.
    
    - Consider text, visuals, context, sarcasm, irony, and intent.
    
    - Explicitly explain your reasoning before giving the final label.
   \vspace{10pt}
    
    Rules:
    
    - The meme may contain English, Hindi, or Chinese text.
    
    - Focus on meaning rather than language.
    
    - Watch for sarcasm, irony, or parody.
    
    - If the meme criticizes or mocks homophobia or transphobia, classify it as Non\_Anti\_LGBT.
    
    - The label should reflect the target and intent of the meme, not just keywords.
    
    - If unsure, choose the closest valid label.
    \vspace{10pt}
    
    Output format (strictly follow this format):
    
    <reason>
    
    Your step-by-step reasoning here.
    
    </reason>
    
    <output>
    
    One label only: Homophobia, Transphobia, or Non\_Anti\_LGBT
    
    </output>

    \vspace{10pt}
    - Do NOT put the label outside the <output> tags.
    
    - Do NOT include anything outside these tags.
    
    - Do NOT include multiple labels.
    
    - Ensure the final answer appears only inside <output> tags."""
    \end{tcolorbox}

    \caption{System prompt baseline classification with allowing external reasoning.}
    \label{fig:basewithsysprompt}
\end{figure}

\section{Feature Extraction Answer-Integer Mapping}
\label{app:answermapping}
The model is instructed to produce a short, constrained answer (e.g., \textit{yes}/\textit{no} or a small set of categorical options). The generated output is then matched against a predefined list of valid answers. If the output does not match any valid option, the query is repeated, with a maximum of ten retries. In the rare case that no valid response is obtained after all retries, a default fallback answer, which is unused elsewhere in the label space, is assigned to ensure completeness.
Table~\ref{tab:answer_mapping} shows the mapping between valid VLM generated answers and a representative integer.

\begin{table}[h]
\centering
\small
\begin{tabular}{p{5.5cm} l}
\toprule
\textbf{Answers} & \textbf{Integer} \\
\midrule

no, No, NO, nah, n, false, False & 0 \\
yes, Yes, YES, yeah, y, true, True & 1 \\
\midrule

0, zero & 0 \\
1, one & 1 \\
2, two & 2 \\
3, three & 3 \\
4, four & 4 \\
5, five & 5 \\
\midrule

A, a, homophobic, Homophobic, gay people & 0 \\
B, b, transphobic, Transphobic, transgender people & 1 \\
C, c, neither, Neither, neutral, none, no group & 2 \\
\midrule

sexual orientation, orientation & 0 \\
gender identity, gender & 1 \\
neither, neutral, none, no target & 2 \\
\midrule

INV (Default) & 6 \\
\bottomrule
\end{tabular}
\caption{Mapping from answer variants to integer representations.}
\label{tab:answer_mapping}
\end{table}

\section{Number of Features After Refinement}
\label{app:featsafterref}

Table~\ref{tab:removedfeats} shows the total number of features after each step of refinements.

\begin{table}[ht]
    \centering
    \setlength{\tabcolsep}{0.8mm}{}
    \renewcommand{\arraystretch}{1.3}
    \begin{tabular}{lcccc}
    \toprule
    Method & English  & Chinese & Hindi & All\\
    \hline
    Base model              & 59 & 59 & 59 & 59 \\
    \textit{AddLF}                   & 89 & 89 & 89 & 89 \\
    \textit{AddLF} + \textit{F1Prune}         & 87 & 88 & 85 & 80 \\
    \textit{AddLF} + \textit{ImpPrune}        & 69 & 38 & 33 & 4 \\
    \bottomrule
    \end{tabular}
    \caption{Number of features considered after each refinement method.}
    \label{tab:removedfeats}
\end{table}

\section{Feature Pattern UMAP Projections}
\label{app:featurepattern}
\begin{figure}[ht]
    \centering
    \includegraphics[width=\linewidth]{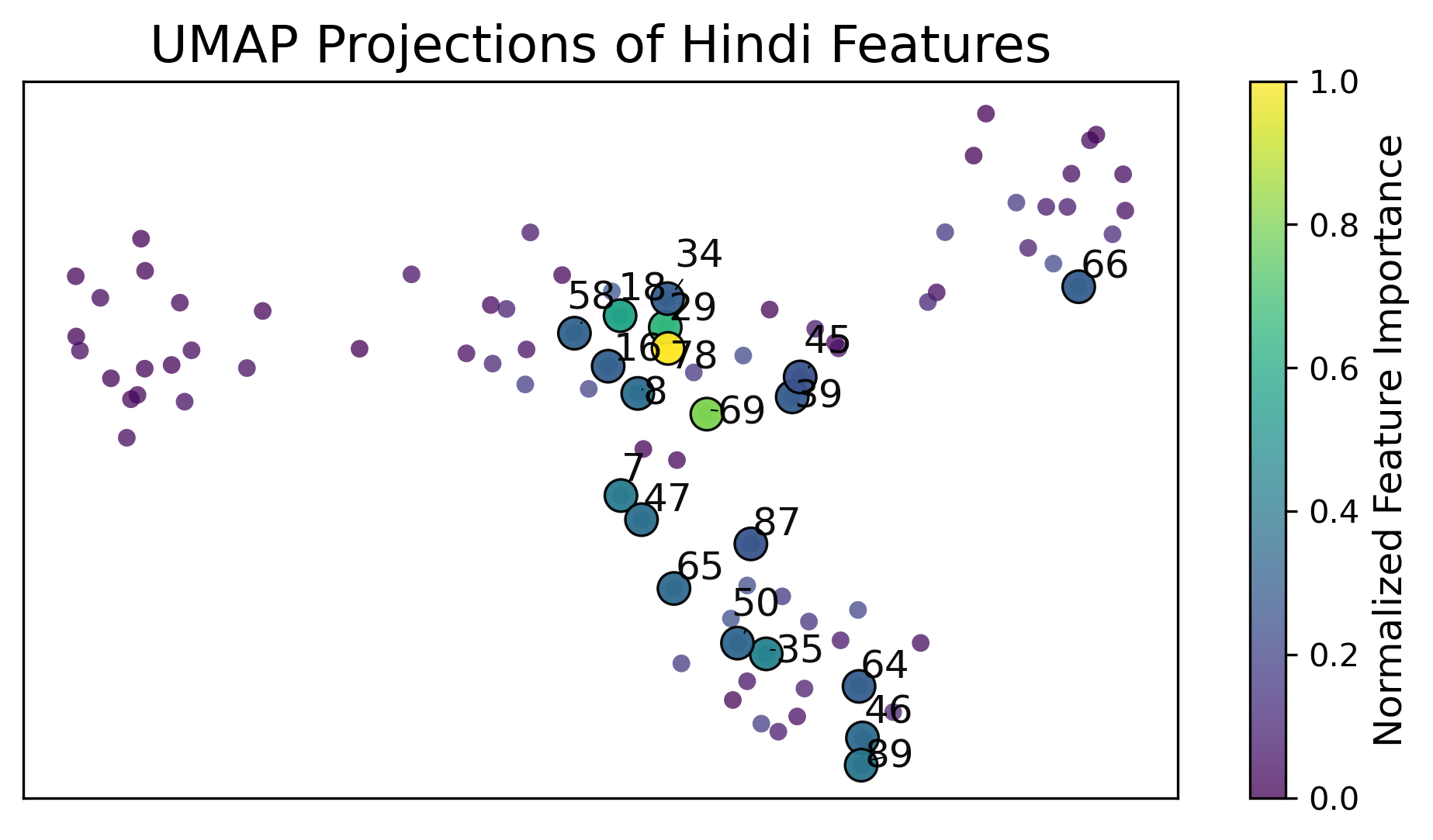}
    \caption{UMAP projection of English question-level features colored by Random Forest importance}
    \label{fig:hindi_umap}
\end{figure}

\begin{figure}[ht]
    \centering
    \includegraphics[width=\linewidth]{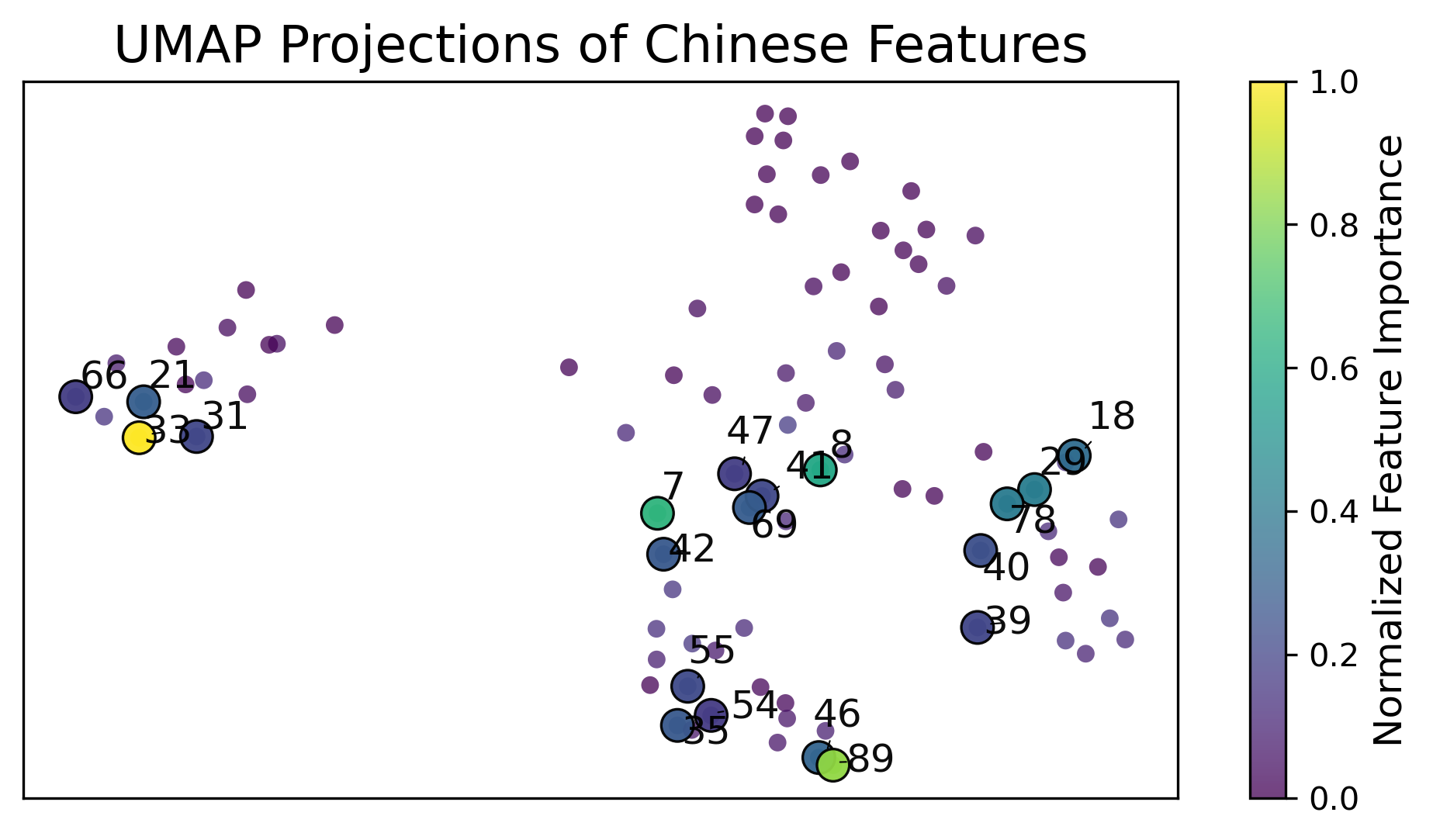}
    \caption{UMAP projection of Chinese question-level features colored by Random Forest importance}
    \label{fig:chinese_umap}
\end{figure}

Figures \ref{fig:hindi_umap} and \ref{fig:chinese_umap} show the UMAP Visualizations of the feature pattern of English and Chinese Memes from the training data respectively. Features in the top 20 highest importances are highlighted.

\end{document}